\documentclass[final]{cvpr}

\usepackage{times}
\usepackage{epsfig}
\usepackage{graphicx}
\usepackage{amsmath}
\usepackage{amssymb}
\usepackage{subfigure}
\usepackage[linesnumbered,boxed]{algorithm2e}
\usepackage{multirow}

\usepackage[pagebackref=true,breaklinks=true,colorlinks,bookmarks=false]{hyperref}

\newcommand\blfootnote[1]{%
  \begingroup
  \renewcommand\thefootnote{}\footnote{#1}%
  \addtocounter{footnote}{-1}%
  \endgroup
}
\def\cvprPaperID{67} 
\def\confYear{CVPR 2021}

\begin{document}

\title{S3Net: A Single Stream Structure for Depth Guided Image Relighting}

\author{Hao-Hsiang Yang$^{1*}$, Wei-Ting Chen$^{1,2*}$, and Sy-Yen Kuo$^{3}$\vspace{1mm}\\
$^{1}$ ASUS Intelligent Cloud Services, Asustek Computer Inc, Taipei, Taiwan\\
$^{2}$ Graduate Institute of Electronics Engineering, National Taiwan University, Taipei, Taiwan\\
$^{3}$ Department of Electrical Engineering, National Taiwan University, Taipei, Taiwan.\\{\tt\small (islike8399, jimmy3505090)@gmail.com, sykuo@ntu.edu.tw}\\ \small\url{https://github.com/dectrfov/NTIRE-2021-Depth-Guided-Image-Any-to-Any-relighting}
}


\maketitle
\blfootnote{*Equally-contributed first authors.}

\begin{abstract}
Depth guided any-to-any image relighting aims to generate a relit image from the original image and corresponding depth maps to match the illumination setting of the given guided image and its depth map. To the best of our knowledge, this task is a new challenge that has not been addressed in the previous literature. To address this issue, we propose a deep learning-based neural \textbf{S}ingle \textbf{S}tream \textbf{S}tructure network called S3Net for depth guided image relighting. This network is an encoder-decoder model. We concatenate all images and corresponding depth maps as the input and feed them into the model. The decoder part contains the attention module and the enhanced module to focus on the relighting-related regions in the guided images. Experiments performed on challenging benchmark show that the proposed model achieves the $3^{rd}$ highest SSIM in the NTIRE 2021 Depth Guided Any-to-any Relighting Challenge.
\end{abstract}

\section{Introduction}
\begin{figure*}[t]
  \centering
\includegraphics[width=\linewidth]{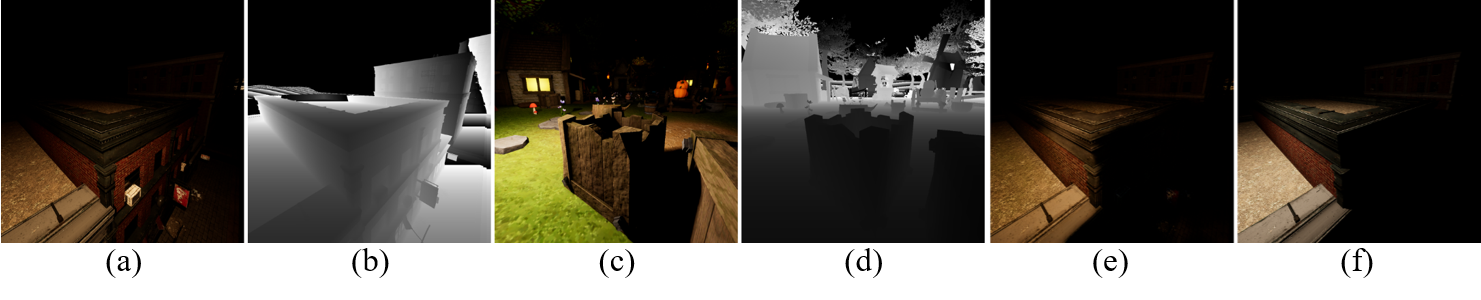}
    \caption{{An example of any-to-any relighting. (a) Original image. (b) Original depth map. (c) Guided image. (d) Guided depth map. (e) Relit image by our method. (f) Ground truth.
}}

\label{fig:example}
\end{figure*}

The objective of this paper is to address the depth guided any-to-any relighting task. That is, an input image with a certain color temperature and light source position setting is relit to match the illumination setting of another guided image. The example is plotted in \figref{fig:example}. \figref{fig:example} (a) is the original source image, and \figref{fig:example} (c) is the guided image. Moreover, the depth maps of two images are provided as shown in \figref{fig:example} (b) and \figref{fig:example} (d), respectively. The target of relighting is to generate a novel result as shown in \figref{fig:example} (f) based on the content of the source image and the light condition of the guided image. \figref{fig:example} (e) is a result estimated by our proposed approach. To the best of our knowledge, this task is a new challenge that has not been addressed in the previous literature. Image relighting is an emerging and crucial technology owing to its applications in visualization, image editing and augmented reality (AR). For example, any-to-any relighting can be used to render images with various ambient lighting conditions for the first and the third person gaming.

The input of any-to-any image relighting is an original image and a guided image, which can be seen as the application of the style transferring \cite{Gong_2018_ECCV}. That is, two images containing different ambient conditions can be seen as different style images. However, there are two inherent differences between the style transferring and the image relighting. First, to generate the image with different ambient conditions, it is necessary to generate shadows into the relit image. Second, similarly, the shadow from the original image needs to be removed. On the other hand, the style transferring focuses on the texture rendering. To address this issue, recently, many deep learning-based image relighting algorithms \cite{puthussery2020wdrn,hu2020sa,guo2019deep,xu2018deep, yang2021multi} are proposed because the deep convolutional neural networks (CNNs) have achieved a lot of successful improvements in many computer vision tasks. They develop the neural networks and follow the end-to-end manner to directly generate relit images without assuming any physical priors. Inspired by those approaches, in this paper, we also propose the deep learning network to tackle the depth guided any-to-any relighting.

Different from the conventional image relighting tasks \cite{elhelou2020aim}, NTIRE 2021 Depth Guided Any-to-any Relighting Challenge provides additional depth maps, which are beneficial for the model to learn spatial representations. To effectively extract the RGB-image and the depth features, the most popular methods are using a dual stream backbone \cite{pang2020hierarchical, chen2020progressively}. However, it is difficult to design the dual backbones to extract the image and depth features because the input information may cause a huge computational burden. Specifically, in NTIRE 2021 Depth Guided Any-to-any Relighting Challenge \cite{elhelou2021ntire}, the sizes of both images and depth maps are 1024 $\times$ 1024. Furthermore, the input contains two images and two depth maps. Therefore, in this paper, we design a single stream structure network (S3Net) to fulfill depth guided any-to-any relighting. This network is an encoder-decoder model. All images and corresponding depth maps are concatenated as the input and passed our network to render the relit image.

To design an efficient network for image relighting, several modules can be utilized. For example, multi-scale feature extractors \cite{qu2019enhanced, yang2021LAFFNet} can be used to increase the receptive field and integrate the coarse-to-fine representation. It is necessary to adopt this module because relit images contain objects in various scales. Additionally, the attention mechanisms are widely used in various tasks like image enhancement \cite{yang2020wavelet, chen2020pmhld} and machine translation \cite{devlin2018bert}. The attention mechanism assigns feature map weights so that features of the sequence of regions or locations are magnified. Because the relit image contains the information of direction (e.g., \figref{fig:example} (a)), the attention mechanism is beneficial for the model to learn the directional representations. Therefore, two modules are selected and integrated into decoder parts to focus on the relighting-related regions in the guided images.
Besides model structure, object functions also impact the overall performance. We combine discrete wavelet transform (DWT) to design a multi-scale loss function to optimize our model so that our model can relight the global ambient conditions and detailed structure.

We summarized the contributions in this paper as follow.
\begin{enumerate}
\item We propose a \textbf{S}ingle \textbf{S}tream \textbf{S}tructure network (S3Net) for the depth guided any-to-any relighting. We apply the single stream network to extract the image and depth features, and deal with various ambient lighting conditions such as the direction of the illumination and the color temperature.
\item During the training phase, we adopt the loss function combining Discrete wavelet transform to train our model. This loss function effectively improves accuracy.
\item We test our proposed method on the VIDIT dataset \cite{helou2020vidit} and multiple experiments demonstrate that the proposed S3Net achieves the $3^{rd}$ highest SSIM and MPS in the NTIRE 2021 Depth Guided Any-to-any Relighting Challenge.
\end{enumerate}
\section{Related Works}
\subsection{RGB-D Fusion}
Different from the conventional RGB tasks, extra depth information can improve accuracy in many computer vision tasks like semantic segmentation \cite{chen2020-SAGate} and object detection \cite{guo2019deep}. Depth maps have demonstrated to be a useful cue to provide geometric and spatial information when combining with the RGB representation. In \cite{guo2019deep}, they propose Faster-RCNN structure to tackle pedestrian detection. They prove that depth maps can be utilized to refine the convolutional features extracted from RGB images. Additionally, more accurate region proposals are achieved by exploring the perspective projection with the help of the depth information. In \cite{chen2020-SAGate}, a unified and efficient cross-modality guided
encoder for semantic segmentation is proposed. This structure jointly filters and recalibrates both representations before cross-modality aggregation. Meanwhile, a bi-direction multi-step propagation strategy is introduced to effectively fuse information between the two modalities. Chen \textit{et.al} \cite{HDFNet-ECCV2020} propose the approach to RGB-D salience object detection. The depth and image features are extracted by a dual backbone. Authors also integrate both features through densely connected structures and apply their mixed features to generate dynamic filters with receptive fields of different sizes. These works tend to use a dual backbone to extract features of different modalities. In practice, the size of relit images in VIDIT \cite{helou2020vidit} and the corresponding depth map is very large (e.g., 1024$\times$1024). The relit images contain illuminant direction information, which is not proper to crop large images to small patches during training. To address this issue, we design a single stream structure to jointly extract both depth and the image features.
\subsection{Deep Learning Based Image Relighting}

Following the rule in the 2021 NTIRE Depth Guided Image Relighting Challenge, there are two kinds of settings. First, the illuminant direction and the temperature are pre-defined \cite{puthussery2020wdrn,yang2021multi}, which is known as one-to-one relighting. Second, the ambient condition is based on a guided image \cite{hu2020sa}, which is known as any-to-any relighting. Both of them are very similar to other low-level vision tasks like image dehazing \cite{chen2020pmhld} image deraining \cite{yang2020wavelet}, image smoke removal \cite{chen2018color}, image desnowing \cite{chen2020jstasr}, reflection removal \cite{tsai2018efficient}, and underwater image enhancement \cite{yang2021LAFFNet}.
All of them belong to the image-to-image translation problems. To tackle those problems, the encoder-decoder network can be applied. Furthermore, when it comes to encoder-decoder structure, U-net \cite{ronneberger2015u,yang2019wavelet} is the most popular network for image-to-image translation tasks. This structure consists not only the encoder-decoder structure but also the skip connection, which concatenates the features with identical size from the encoder to the decoder. For example, Puthussery \textit{et.al} \cite{puthussery2020wdrn} proposes a wavelet decomposed relit network for image relighting. This structure is a novel encoder-decoder network employing wavelet-based decomposition followed by convolution layers under a multi-resolution framework. Additionally, this network contains skip connection operations. Different from those low-level vision tasks, the relit image contains the information of illuminant direction and color temperature, which may cause the shadow of the objects to be located in different locations. In \cite{hu2020sa}, a lighting-to-feature network is proposed to recover the corresponding implicit scene representation from the illumination settings, which is known as the inverse process of the lighting estimation network.

\subsection{Wavelet Transform for Deep Learning}
DWT \cite{mallat1999wavelet} with orthogonal property is widely adopted in many computer vision tasks \cite{yang2019wavelet,puthussery2020wdrn}. DWT decomposes an image into various small patches in different frequency intervals, which can replace the existing down-sampling operations like max pooling or average-pooling. Therefore, many tasks apply DWT to diminish feature maps and achieve multi-scale features. Moreover, in \cite{yang2020net}, DWT is leveraged to design the objective function to measure the similarity between the ground truth and predict images. Motivated by these works, we also combine DWT in the loss function to make our network learn the multi-scale representations. 

\subsection{Attention Mechanism}
Attention mechanisms are important roles in both human perception systems and deep learning tasks \cite{yang2020wavelet,mnih2014recurrent}. Attention mechanisms provide feature maps or certain sequence weights so that features of regions or locations can be magnified. Specifically, for computer vision tasks, attention mechanisms are categorized as the spatial attention \cite{hu2020sa} and the channel attention \cite{yang2020wavelet}. The former spatially utilizes weights to refine the feature maps, and the latter computes the global average-pooled features to implement the channel-wise attention. In this paper, both attention mechanisms are leveraged in our model to further increase the performance of the network. 

\section{Proposed Methods}

\begin{figure*}[h]
  \centering
\includegraphics[width=0.7\textwidth ]{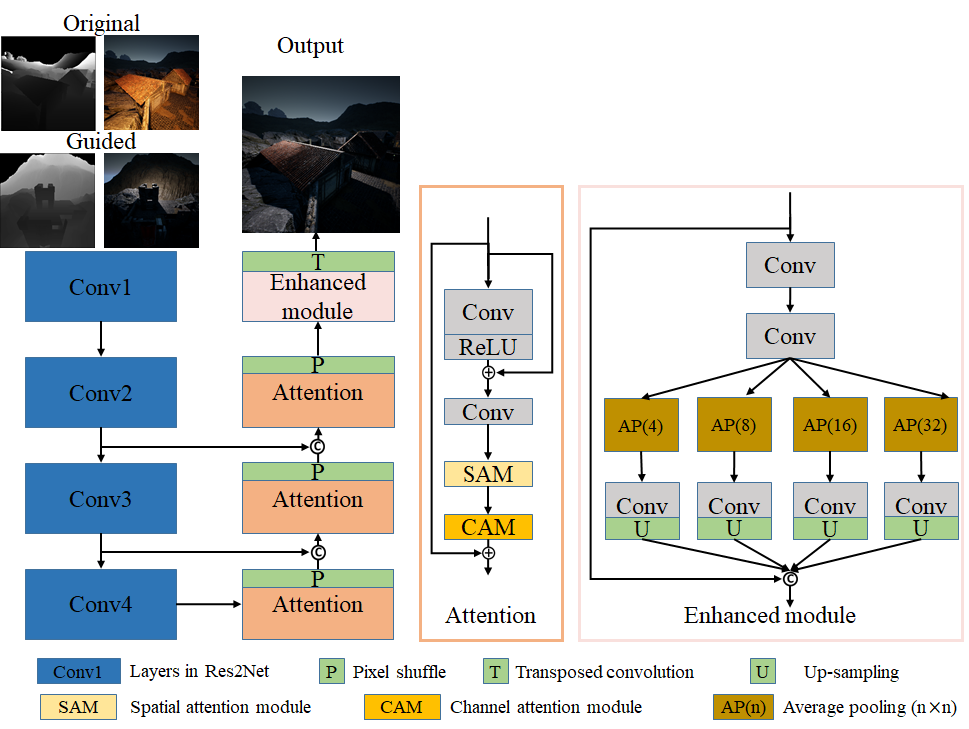}

    \caption{{The proposed network for depth guided any-to-any relighting. This model applies the Res2Net as the encoder to extract both the image and depth map features. All images and depth maps are concentrated as input. In the decoder parts, we use attention mechanism including channel and spatial attention mechanism and the enhanced module to refine features and relight the image.
\label{fig:architecture}
}}

\label{fig:figure2}
\end{figure*}

\subsection{Overall Neural Network}
 The architecture of the proposed S3Net is presented in \figref{fig:architecture}. This network is based on \cite{wu2020knowledge} and contains the encoder and decoder parts. In the encoder part, we apply the Res2Net101 \cite{gao2019res2net} as our backbone. The Res2Net can represent multi-scale features at a granular level and increases the range of receptive fields for each network layer. After the input is passed through the backbone, multi-scale feature extraction is achieved. Note that, the Res2Net utilized in this work discards the full connection layer and the size of final output feature maps from our encoder is $\frac{1}{16}$. The initial weight of the encoder is pre-trained parameters trained from the ImageNet. We connect bottom features to the decoder. The decoder consists of the stacks of convolution to refine the feature maps. Both the pixel shuffle \cite{shi2016real} and the transposed convolution \cite{gao2019pixel} are adopted to magnify feature maps. Furthermore, we leverage attention modules to refine intermediate features. Attention modules consist of residual layer \cite{he2016deep}, the spatial \cite{mnih2014recurrent}, and the channel attention \cite{hu2018squeeze} as shown in \figref{fig:architecture} (orange rectangle).

Furthermore, inspired by \cite{qu2019enhanced}, we add the enhanced module in our S3Net. The enhanced module leverages average pooling in different strides to change the size of feature maps and receptive fields, which is effective to extract multi-scale features. Finally, the up-sampling is applied to restore diminished feature maps and all feature maps are concatenated. The enhanced module is illustrated in the light orange rectangle in \figref{fig:architecture}. Moreover, it is known that U-Net-like structure is beneficial in many tasks such as image dehazing \cite{chen2019pms,chen2020pmhld} and semantic segmentation \cite{ronneberger2015u}. Its skip connection encourages feature re-using. Therefore, in our S3Net, we also adopt skip connection to merge the last three feature maps from the backbone to their corresponding feature maps. 

In practice, we directly combine the original image, original depth map, guided image and guided depth map as input. This input is seen as the 8-channel tensor and the output is the 3-channel relit image. In addition, we change the first convolution in the Res2Net backbone so that the S3Net can accommodate 8-channel tensors as input. 

\subsection{Loss Functions}
To train our S3Net, we apply three loss functions. The first function is Charbonnier loss \cite{barron2019general} that can be regarded as the robust $L_1$ loss function. The Charbonnier loss is expressed as:
\begin{equation}
L_{Cha}(I, \hat I) = \frac{1}{T}\sum_{i}^{T}\sqrt[]{(I_i-\hat{I_i})^2+\epsilon ^2}
\label{eq:res}
\end{equation}
where $I$ and $\hat I$ represent the ground truth and relit images from the proposed network, respectively, and $e$ is seen as a tiny constant (e.g., $10^{-6}$) for stable and robust convergence. $L_Cha$ can restore global structure \cite{barron2019general} and can be more robust to handle outliers.

Secondly, we apply the SSIM loss \cite{zhao2016loss}. The SSIM loss is able to reconstruct local textures and details. It can be expressed as:
\begin{equation}
  {L_{SSIM}}(I,\hat I) = -\frac{(2\mu_I\mu_{\hat I} + C_1)  (2 \sigma _{I \hat I} + C_2)} 
    {(\mu_I^2 + \mu_{\hat I}^2+C_1) (\sigma_I^2 + \sigma_{\hat I}^2+C_2) }
  \label{eq:SSMI}
\end{equation}
where {$\sigma$} and {$\mu$} represent the standard deviation, the covariance and the mean of images. In the image relighting task, to remove shadows from the original image, we extend the SSIM loss function so that our network can restore more detailed parts. We follow the method in \cite{yang2020net} and combine DWT into the SSIM loss because lots of tasks \cite{yang2020wavelet,yang2019wavelet} have demonstrated that the DWT captures the high-frequency features, which is beneficial for reconstructing the clear details on relit images. Initially, the DWT decomposes the predicted image into four different and small sub-band images. The operation can be expressed as:
\begin{equation}
\hat I^{LL}, \hat I^{LH}, \hat I^{HL}, \hat I^{HH} = {\rm DWT}(\hat I)
\label{eq:gan}
\end{equation}
where superscripts mean the output from respective filters (e.g., {$f_{LL}$}, {$f_{HL}$}, {$f_{LH}$} and {$f_{HH}$}). 

$f_{HL}$ ,$f_{LH}$ and $f_{HH}$ are high-pass filters for the horizontal edge, the vertical edge and the corner detection, respectively. $f_{LL}$ is seen as the down-sampling operation.
Moreover, the DWT can keep decomposing the $\hat I_{LL}$ to generate images with different scales and frequency information. This step is written as:
\begin{equation}
\hat I^{LL}_{i+1}, \hat I_{i+1}^{LH}, \hat I_{i+1}^{HL}, \hat I_{i+1}^{HH} = {\rm DWT}(\hat I_{i}^{LL})
\label{eq:gan}
\end{equation}
where the subscript $i$ means the output from the $i^{th}$ DWT iteration, and $\hat I^{LL}_{0}$ is the original predicted relit image. The SSIM loss terms described above are calculated from the original image pair and various sub-band image pairs. The fusion of the SSIM loss and the DWT is integrated as:
\begin{equation}
\begin{aligned}
  &L_{{W-SSIM}}(I,\hat I) = \sum_{0}^{i} \gamma_{i}L_{{\rm SSIM}}(I^{w}_i,\hat I^{w}_i), \\
&w \in \left\{ LL, HL, LH, HH\right\}
\end{aligned}
\label{eq:SSMIloss}
\end{equation}
where $\gamma_{i}$ is based on \cite{yang2020net} to control the importance of different patches.

The third loss is the perceptual loss \cite{johnson2016perceptual}. Different from the aforementioned two loss functions, the perceptual loss leverages multi-scale features achieved from a pre-trained deep neural network (e.g., VGG19 \cite{simonyan2014very}) to measure the visual feature difference between the ground truth and the estimated image. Formally, in this task, the VGG19 pre-trained on ImageNet is utilized as the loss function network. The perceptual loss is defined as

\begin{equation}
L_{Per}(I,\hat{I})= |(VGG(I)-VGG(\hat{I})|
\label{eq:res}
\end{equation}
where $| \cdot |$ is the absolute value. The overall loss function is expressed as:
\begin{equation}
L_{Total}= \lambda_{1} L_{cha}+ \lambda_{2} L_{W-SSIM} + \lambda_{3} L_{Per}
\label{loss}
\end{equation}
where $\lambda_{1}$, $\lambda_{2}$  and $\lambda_{3}$ are scaling coefficients and used to adjust the relative weights on the three components.

\section{Experiments}
\subsection{Dataset}
The dataset used in the 2021 NTIRE image challenge of depth guided image relighting is the Virtual Image Dataset for Illumination Transfer (VIDIT) \cite{helou2020vidit}. This dataset consists of 390 different scenes which are with 40 different illumination settings (five different color temperatures from 2500 to 6500K and 8 azimuthal angles). There are 15600 images totally. Additionally, the corresponding 390 depth maps are provided. The size of all training images and depth maps are 1024 × 1024 × 3 and 1024 × 1024 × 1, respectively. 300 virtual scenes with different illumination settings are applied for training while the other 90 virtual scenes are used for validation and testing.

\subsection{Experimental Setting}

\begin{figure*}[t]
  \centering
\includegraphics[width=0.98\linewidth]{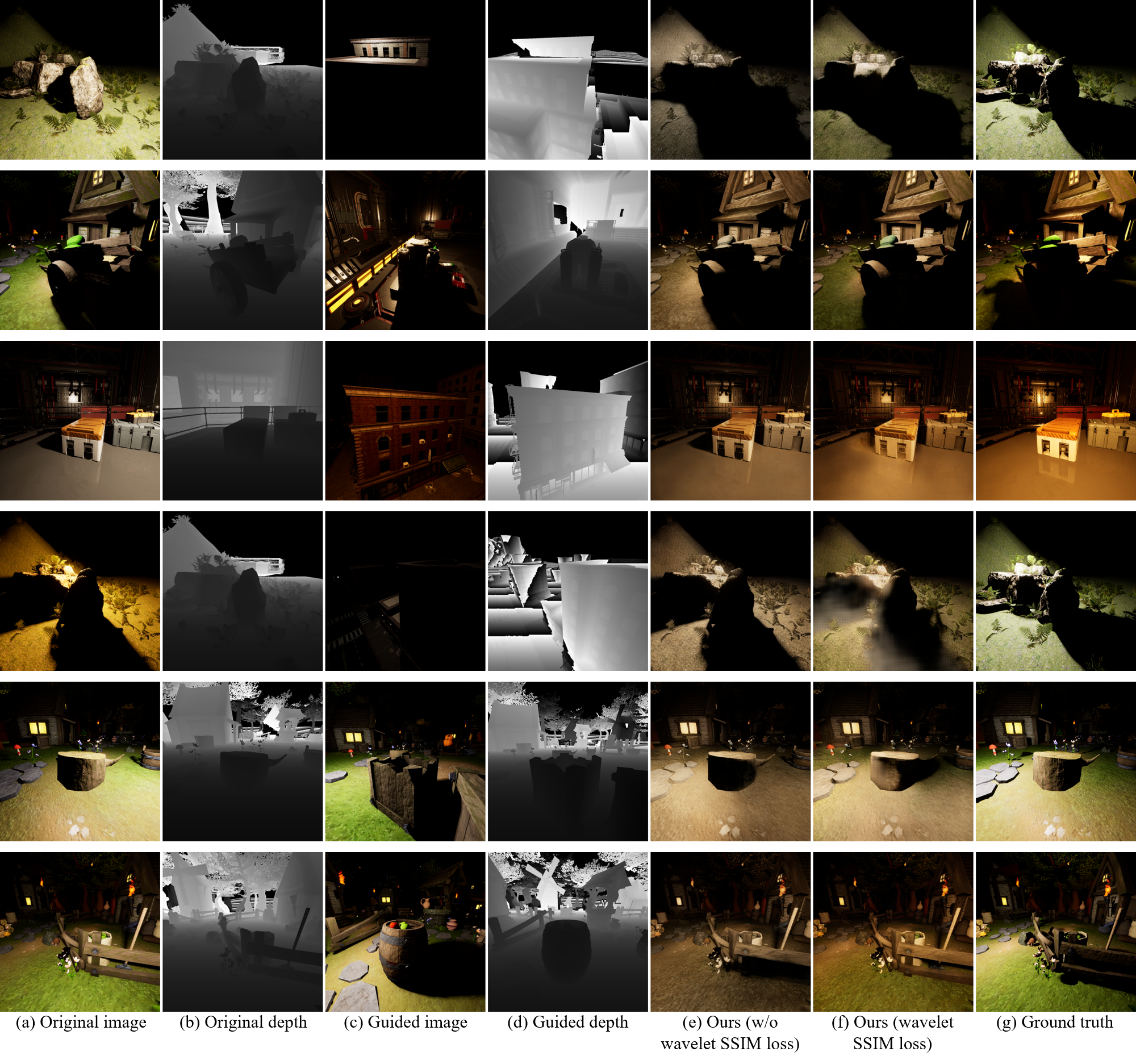}
    \caption{{The visual comparison of the proposed method and other existing methods. We plot relit results using the proposed S3Net trained with and without the wavelet SSIM loss. 
}}

\label{fig:vis}
\end{figure*}

In the training phase, we first choose an arbitrary image and the corresponding depth map. Then, we choose an image as a guided image that contains random color temperature and azimuthal angles. With this information, the output image can be determined. Our network utilizes the original image, the original depth map, the guided image and the guided depth map as input to reconstruct the final decided image. We do not apply any data augmentation like random flip and random cropping during the training phase.

we apply the AdamW \cite{loshchilov2017decoupled} as the new optimizer, and the batch size is set as 3 and the network is trained for 100 epochs with the momentum $\beta_{1}$ = 0.5 and $\beta_{2}$ = 0.999. The learning rate is set as $10^{-4}$ and it is divided by ten after 20 epochs. The weights of these three loss terms (i.e., $\lambda_{1}, \lambda_{2}$ and $\lambda_{3}$) are set as 1, 1.1 and 0.1, respectively. All experiments are performed on the PyTorch platform and a single Nvidia V100 graphic card. We spend about 80 hours finishing the model training.\footnote{The source code will be released in our project page.}

\subsection{Ablation Experiments}

\begin{figure*}[t]
  \centering
\includegraphics[width=0.92\linewidth]{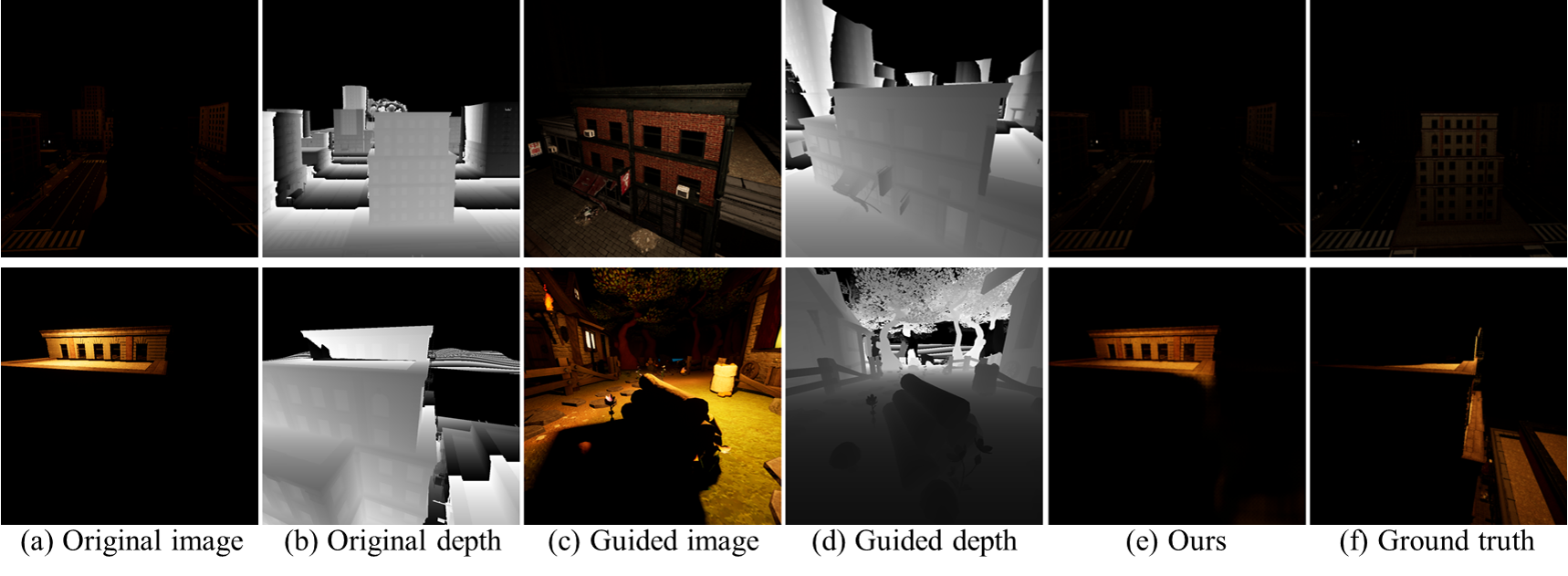}
    \caption{{The visualization of the failure cases recovered by the proposed S3Net. One see that, though the color temperature can be transferred successfully, the detailed and structural information under the regions of shadows cannot be recovered adequately.}}

\label{fig:fail}
\end{figure*}

To find the best effectiveness of the proposed S3Net, some ablation experiments are conducted in this section. The peak signal-to-noise ratio (PSNR) and the structural similarity (SSIM) are adopted as objective metrics for quantitative evaluation. 

The ablation experiments consist of three experimental settings: 1) The input is only with the source image and guided image but without depth maps, that is, the input is a six-channel tensor (Image); 2) The input is with the source image, guided image, and their corresponding depth maps (Both); For these two settings, $L_{Cha}$, $L_{SSIM}$ and $L_{Per}$ are adopted as objective functions. 3) The same setting in 2) but the $L_{SSIM}$ \cite{yang2020net} is replaced by $L_{W-SSIM}$. The results are reported in Table \ref{tab:ablation1}. One can see that the PSNR and SSIM scores of setting 2 can be improved compared with setting 1. It can prove that using depth maps can improve the performance of relighting because they can provide more spatial information for the network. Moreover, compared with setting 2, the performance of setting 3 is improved effectively. It indicates that the $L_{W-SSIM}$ in the network can be more beneficial compared with the $L_{SSIM}$.

\begin{table}[h]
    \centering
    \caption{{The comparison of using different input data and applying the different loss functions on VIDIT dataset.}}
\begin{tabular}{l|cccc}
\hline
Index     & Input &$L_{W-SSIM}$ & PSNR & SSIM \\ \hline\hline
1 & Image&  & 18.7611    & 0.6821 \\
2 & Both &    & 18.8451     &  0.6913    \\
3&  Both&  $\surd$ & {\color{red}19.1281}   &  {\color{red}0.6969}
\end{tabular}
\label{tab:ablation1}
\end{table}

\begin{table*}[t!]
  \centering
  \caption{The average SSIM, PSNR, MPS and LPIPS  of top five methods over NTIRE 2021 depth guided image relighting validation and testing dataset.}
    \begin{tabular}{r|cc|cccc}
    \multicolumn{1}{l|}{User name} & \multicolumn{2}{c|}{Validation} & \multicolumn{4}{c}{Testing} \\

          & SSIM  & PSNR  & MPS   & SSIM  & LPIPS & PSNR \\ \hline
    DeepBlueAI &0.7196&20.0637& 0.7675 & 0.7087 & 0.1737 & 20.7915 \\
    lifu & 0.7107&19.7730&0.7671 & 0.6874 & 0.1532 & 19.8901 \\
    elientumba &0.6802&18.5570& 0.7423 & 0.6508 & 0.1661 & 18.6039 \\
    auy200& 0.6864&19.3552&0.7341 & 0.6711 & 0.2028 & 20.1478 \\ \hline \hline
    HaoqiangYang &0.7022&19.2462 & 0.6452&0.6784 & 0.1566 & 19.2212 \\
      
    \end{tabular}%
  \label{tab:addlabel}%
\end{table*}%




We also present some visual comparisons from VIDIT validation set for depth guided any-to-any relighting problem to prove the effectiveness of using the wavelet SSIM loss in the proposed S3Net. As shown in \figref{fig:vis}, images relit from the model trained with wavelet SSIM loss are closer to the ground truth. This model can properly remove shadows and transfer the ambient conditions from guide images to source images.


\subsection{Results of Challenge}



We list the results of the proposed S3Net compared with other competing entries in depth guided any-to-any relighting challenge of NTIRE 2021 workshop \cite{elhelou2021ntire} in Table \ref{tab:addlabel}. Besides PSNR and SSIM, the Mean Perceptual Score (MPS) defined as the average of the normalized SSIM and LPIPS \cite{zhang2018perceptual} scores, themselves averaged
across the entire test set of each submission is adopted to evaluate performance of all submissions. 
As shown in Table \ref{tab:addlabel}, our results obtained the $3^{rd}$ place, the $4^{th}$ place, the $2^{nd}$ place and the $3^{rd}$ place in terms of SSIM, PSNR, LPIPS and MPS. It is noted that our model takes 2.042 seconds on average to generate a relit result with 1024 $\times$ 1024 image size in the test phase.

\subsection{Limitations and Discussion}
The proposed method learns the mapping functions of depth guided image relighting. Although our method achieves competitive performance in this competition, it may fail under the certain conditions. As shown in \figref{fig:fail}, the relit images are very different from the ground truth. Because when original images contain a large region of shadows, our model cannot identify the foreground and the backward of them. Even the depth maps are given, this information just provides the front side instead of the omnidirectional spatial information. Therefore, the color temperature from the guided images is transferred to the relit image (e.g., \figref{fig:fail} (e)), but the model reconstructs the poor structures.
\section{Conclusion}
In this paper, we propose a single stream structure network (S3Net) for depth guided any-to-any image relighting. We concatenate source image and guided image with their corresponding depth maps as input to design our model. This network is based on the Res2Net \cite{gao2019res2net} in the encoder part. The attention modules and the enhanced module \cite{qu2019enhanced} are applied in the decoder part to refine the feature maps. We leverage the wavelet SSIM loss \cite{yang2020net} to supervise the network training. Moreover, in the NTIRE 2021 Depth Guided Any-to-any Image Relighting Challenge, the proposed S3Net achieves the $3^{rd}$ place in terms of PMS and SSIM. Since depth guided image relighting is a new challenge that has not been addressed in the previous literature, in future works, we will design the novel backbone to extract and fuse the image and depth features effectively.
\section{Acknowledgement}
We thank to National Center for High-performance Computing (NCHC) for providing computational and storage resources.
{\small
\bibliographystyle{IEEEtran}
\bibliography{egbib}

\begin{thebibliography}{10}
\providecommand{\url}[1]{#1}
\csname url@samestyle\endcsname
\providecommand{\newblock}{\relax}
\providecommand{\bibinfo}[2]{#2}
\providecommand{\BIBentrySTDinterwordspacing}{\spaceskip=0pt\relax}
\providecommand{\BIBentryALTinterwordstretchfactor}{4}
\providecommand{\BIBentryALTinterwordspacing}{\spaceskip=\fontdimen2\font plus
\BIBentryALTinterwordstretchfactor\fontdimen3\font minus
  \fontdimen4\font\relax}
\providecommand{\BIBforeignlanguage}[2]{{%
\expandafter\ifx\csname l@#1\endcsname\relax
\typeout{** WARNING: IEEEtran.bst: No hyphenation pattern has been}%
\typeout{** loaded for the language `#1'. Using the pattern for}%
\typeout{** the default language instead.}%
\else
\language=\csname l@#1\endcsname
\fi
#2}}
\providecommand{\BIBdecl}{\relax}
\BIBdecl

\bibitem{Gong_2018_ECCV}
X.~Gong, H.~Huang, L.~Ma, F.~Shen, W.~Liu, and T.~Zhang, ``Neural stereoscopic
  image style transfer,'' in \emph{Proceedings of the European Conference on
  Computer Vision (ECCV)}, 2018.

\bibitem{puthussery2020wdrn}
D.~Puthussery, M.~Kuriakose, J.~C~V \emph{et~al.}, ``{WDRN}: A wavelet
  decomposed relightnet for image relighting,'' \emph{arXiv preprint
  arXiv:2009.06678}, 2020.

\bibitem{hu2020sa}
Z.~Hu, X.~Huang, Y.~Li, and Q.~Wang, ``{SA-AE} for any-to-any relighting,'' in
  \emph{European Conference on Computer Vision}, 2020.

\bibitem{guo2019deep}
Z.~Guo, W.~Liao, Y.~Xiao, P.~Veelaert, and W.~Philips, ``Deep learning fusion
  of rgb and depth images for pedestrian detection,'' in \emph{British Machine
  Vision Conference}, 2019.

\bibitem{xu2018deep}
Z.~Xu, K.~Sunkavalli, S.~Hadap, and R.~Ramamoorthi, ``Deep image-based
  relighting from optimal sparse samples,'' \emph{ACM Transactions on Graphics
  (ToG)}, 2018.

\bibitem{yang2021multi}
H.-H. Yang, W.-T. Chen, H.-L. Luo, and S.-Y. Kuo, ``Multi-modal bifurcated
  network for depth guided image relighting,'' in \emph{Proceedings of the IEEE
  Conference on Computer Vision and Pattern Recognition Workshops (CVPRW)},
  2021.

\bibitem{elhelou2020aim}
M.~El~Helou, R.~Zhou, S.~S\"usstrunk, R.~Timofte \emph{et~al.}, ``{AIM} 2020:
  Scene relighting and illumination estimation challenge,'' in
  \emph{Proceedings of the European Conference on Computer Vision Workshops
  (ECCVW)}, 2020.

\bibitem{pang2020hierarchical}
Y.~Pang, L.~Zhang, X.~Zhao, and H.~Lu, ``Hierarchical dynamic filtering network
  for rgb-d salient object detection,'' \emph{arXiv preprint arXiv:2007.06227},
  2020.

\bibitem{chen2020progressively}
S.~Chen and Y.~Fu, ``Progressively guided alternate refinement network for
  rgb-d salient object detection,'' in \emph{European Conference on Computer
  Vision}, 2020.

\bibitem{elhelou2021ntire}
M.~El~Helou, R.~Zhou, S.~S\"usstrunk, R.~Timofte \emph{et~al.}, ``{NTIRE} 2021:
  Depth-guided image relighting challenge,'' in \emph{Proceedings of the IEEE
  Conference on Computer Vision and Pattern Recognition Workshops (CVPRW)},
  2021.

\bibitem{qu2019enhanced}
Y.~Qu, Y.~Chen, J.~Huang, and Y.~Xie, ``Enhanced pix2pix dehazing network,'' in
  \emph{Proceedings of the IEEE/CVF Conference on Computer Vision and Pattern
  Recognition}, 2019.

\bibitem{yang2021LAFFNet}
H.-H. Yang, K.-C. Huang, and W.-T. Chen, ``Laffnet: A lightweight adaptive
  feature fusion network for underwater image enhancement,'' in \emph{IEEE
  International Conference on Robotics and Automation (ICRA)}, 2021.

\bibitem{yang2020wavelet}
H.-H. Yang, C.-H.~H. Yang, and Y.-C.~F. Wang, ``Wavelet channel attention
  module with a fusion network for single image deraining,'' in \emph{IEEE
  International Conference on Image Processing (ICIP)}, 2020.

\bibitem{chen2020pmhld}
W.-T. Chen, H.-Y. Fang, J.-J. Ding, and S.-Y. Kuo, ``{PMHLD}: patch map-based
  hybrid learning dehazenet for single image haze removal,'' \emph{IEEE
  Transactions on Image Processing}, 2020.

\bibitem{devlin2018bert}
J.~Devlin, M.-W. Chang, K.~Lee, and K.~Toutanova, ``Bert: Pre-training of deep
  bidirectional transformers for language understanding,'' \emph{arXiv preprint
  arXiv:1810.04805}, 2018.

\bibitem{helou2020vidit}
M.~E. Helou, R.~Zhou, J.~Barthas, and S.~S{\"u}sstrunk, ``Vidit: Virtual image
  dataset for illumination transfer,'' \emph{arXiv preprint arXiv:2005.05460},
  2020.

\bibitem{chen2020-SAGate}
X.~Chen, K.-Y. Lin, J.~Wang, W.~Wu, C.~Qian, H.~Li, and G.~Zeng,
  ``Bi-directional cross-modality feature propagation with
  separation-and-aggregation gate for rgb-d semantic segmentation,'' in
  \emph{European Conference on Computer Vision (ECCV)}, 2020.

\bibitem{HDFNet-ECCV2020}
Y.~Pang, L.~Zhang, X.~Zhao, and H.~Lu, ``Hierarchical dynamic filtering network
  for rgb-d salient object detection,'' in \emph{Eur. Conf. Comput. Vis.},
  2020.

\bibitem{chen2018color}
W.-T. Chen, S.-Y. Yuan, G.-C. Tsai, H.-C. Wang, and S.-Y. Kuo, ``Color
  channel-based smoke removal algorithm using machine learning for static
  images,'' in \emph{IEEE International Conference on Image Processing (ICIP)},
  2018.

\bibitem{chen2020jstasr}
W.-T. Chen, H.-Y. Fang, J.-J. Ding, C.-C. Tsai, and S.-Y. Kuo, ``{JSTASR}:
  Joint size and transparency-aware snow removal algorithm based on modified
  partial convolution and veiling effect removal,'' in \emph{European
  Conference on Computer Vision}, 2020.

\bibitem{tsai2018efficient}
G.-C. Tsai, W.-T. Chen, S.-Y. Yuan, and S.-Y. Kuo, ``Efficient reflection
  removal algorithm for single image by pixel compensation and detail
  reconstruction,'' in \emph{IEEE International Conference on Digital Signal
  Processing (DSP)}, 2018.

\bibitem{ronneberger2015u}
O.~Ronneberger, P.~Fischer, and T.~Brox, ``U-net: Convolutional networks for
  biomedical image segmentation,'' in \emph{International Conference on Medical
  image computing and computer-assisted intervention}, 2015.

\bibitem{yang2019wavelet}
H.-H. Yang and Y.~Fu, ``Wavelet {U}-net and the chromatic adaptation transform
  for single image dehazing,'' in \emph{IEEE International Conference on Image
  Processing (ICIP)}, 2019.

\bibitem{mallat1999wavelet}
S.~Mallat, \emph{A wavelet tour of signal processing}.\hskip 1em plus 0.5em
  minus 0.4em\relax Elsevier, 1999.

\bibitem{yang2020net}
H.-H. Yang, C.-H.~H. Yang, and Y.-C.~J. Tsai, ``Y-net: Multi-scale feature
  aggregation network with wavelet structure similarity loss function for
  single image dehazing,'' in \emph{IEEE International Conference on Acoustics,
  Speech and Signal Processing (ICASSP)}, 2020.

\bibitem{mnih2014recurrent}
V.~Mnih, N.~Heess \emph{et~al.}, ``Recurrent models of visual attention,'' in
  \emph{Advances in neural information processing systems}, 2014.

\bibitem{wu2020knowledge}
H.~Wu, J.~Liu, Y.~Xie, Y.~Qu, and L.~Ma, ``Knowledge transfer dehazing network
  for nonhomogeneous dehazing,'' in \emph{Proceedings of the IEEE/CVF
  Conference on Computer Vision and Pattern Recognition Workshops}, 2020.

\bibitem{gao2019res2net}
S.~Gao, M.-M. Cheng, K.~Zhao, X.-Y. Zhang, M.-H. Yang, and P.~H. Torr,
  ``Res2net: A new multi-scale backbone architecture,'' \emph{IEEE transactions
  on pattern analysis and machine intelligence}, 2019.

\bibitem{shi2016real}
W.~Shi, J.~Caballero, F.~Husz{\'a}r, J.~Totz, A.~P. Aitken, R.~Bishop,
  D.~Rueckert, and Z.~Wang, ``Real-time single image and video super-resolution
  using an efficient sub-pixel convolutional neural network,'' in
  \emph{Proceedings of the IEEE conference on computer vision and pattern
  recognition}, 2016.

\bibitem{gao2019pixel}
H.~Gao, H.~Yuan, Z.~Wang, and S.~Ji, ``Pixel transposed convolutional
  networks,'' \emph{IEEE transactions on pattern analysis and machine
  intelligence}, 2019.

\bibitem{he2016deep}
K.~He, X.~Zhang, S.~Ren, and J.~Sun, ``Deep residual learning for image
  recognition,'' in \emph{Proceedings of the IEEE conference on computer vision
  and pattern recognition}, 2016.

\bibitem{hu2018squeeze}
J.~Hu, L.~Shen, and G.~Sun, ``Squeeze-and-excitation networks,'' in
  \emph{Proceedings of the IEEE conference on computer vision and pattern
  recognition}, 2018.

\bibitem{chen2019pms}
W.-T. Chen, J.-J. Ding, and S.-Y. Kuo, ``{PMS}-net: Robust haze removal based
  on patch map for single images,'' in \emph{Proceedings of the IEEE/CVF
  Conference on Computer Vision and Pattern Recognition}, 2019.

\bibitem{barron2019general}
J.~T. Barron, ``A general and adaptive robust loss function,'' in
  \emph{Proceedings of the IEEE Conference on Computer Vision and Pattern
  Recognition}, 2019.

\bibitem{zhao2016loss}
H.~Zhao, O.~Gallo, I.~Frosio, and J.~Kautz, ``Loss functions for image
  restoration with neural networks,'' \emph{IEEE Transactions on Computational
  Imaging}, 2016.

\bibitem{johnson2016perceptual}
J.~Johnson, A.~Alahi, and L.~Fei-Fei, ``Perceptual losses for real-time style
  transfer and super-resolution,'' in \emph{European conference on computer
  vision}, 2016.

\bibitem{simonyan2014very}
K.~Simonyan and A.~Zisserman, ``{Very deep convolutional networks for
  large-scale image recognition},'' \emph{arXiv preprint arXiv:1409.1556},
  2014.

\bibitem{loshchilov2017decoupled}
I.~Loshchilov and F.~Hutter, ``Decoupled weight decay regularization,''
  \emph{arXiv preprint arXiv:1711.05101}, 2017.

\bibitem{zhang2018perceptual}
R.~Zhang, P.~Isola, A.~A. Efros, E.~Shechtman, and O.~Wang, ``The unreasonable
  effectiveness of deep features as a perceptual metric,'' in \emph{CVPR},
  2018.

\end{thebibliography}
}

\end{document}